\documentclass[lettersize,journal]{IEEEtran}
\usepackage{amsmath,amsfonts}
\usepackage{algorithmic}
\usepackage{algorithm}
\usepackage{array}
\usepackage[caption=false,font=normalsize,labelfont=sf,textfont=sf]{subfig}
\usepackage{textcomp}
\usepackage{stfloats}
\usepackage{url}
\usepackage{xcolor}
\usepackage{verbatim}
\usepackage{graphicx}
\usepackage{cite}
\usepackage{tabularx}
\usepackage{booktabs}
\usepackage{multirow}
\usepackage{float}
\usepackage{makecell}
\usepackage{enumitem}
\usepackage{listings}
\begin{document}

\title{PsycoLLM: Enhancing LLM for Psychological Understanding and Evaluation}

\author{Jinpeng Hu,~Tengteng Dong,~Luo Gang,~Hui Ma,~Peng Zou,~Xiao Sun,~Dan Guo,~Xun Yang,\\~Meng Wang\textsuperscript{$\ast$},~\IEEEmembership{Fellow,~IEEE}
\thanks{This work was supported in part by National Natural Science Foundation of China under Grant 62402158, Grant 72188101 and Grant U22A2094; by the Major Project of Anhui Province Grant 202203a05020011, Grant 202423k09020001, and Grant 2408085J040; and by the Fundamental Research Funds for the Central Universities Grant JZ2024HGTG0309, Grant JZ2024AHST0337, Grant JZ2023YQTD0072.
\emph{(*Corresponding authors: Meng Wang.)}}
\thanks{Jinpeng Hu, Tengteng Dong and Hui Ma are with the School of Computer Science and Information Engineering, Hefei University of Technology (HFUT), Hefei 230601, China (e-mail: jinpenghu@hfut.edu.cn; 2022111070@mail.hfut.edu.cn; huima@hfut.edu.cn).
}
\thanks{Luo Gang is with the Key Laboratory of Brain Health Intelligent Evaluation and Intervention, Ministry of Education, Beijing Institute of Technology, Beijing 10083, P. R. China and also with the School of Medical Technology, Beijing Institute of Technology, Beijing 10083, P. R. China (e-mail: gang@bit.edu.cn).
}
\thanks{Peng Zou is with the Institute of Artificial Intelligence, Hefei Comprehensive National Science Center, Hefei 230026, China (e-mail: zonepg@mail.ustc.edu.cn).
}
\thanks{Xiao Sun is with the School of Computer Science and Information Engineering, Hefei University of Technology (HFUT), Hefei 230601, China, also with the Institute of Artificial Intelligence, Hefei Comprehensive National Science Center, Hefei 230026, China and also with the Hefei Zhongjuyuan Intelligent Technology Co., Ltd., Hefei 230088, China (e-mail: sunx@hfut.edu.cn).
}
\thanks{Xun Yang is with the School of Information Science and Technology, University of Science and Technology of China (USTC), Hefei 230026, China (e-mail: xyang21@ustc.edu.cn).
}
\thanks{Dan Guo and Meng Wang are with the School of Computer Science and Information Engineering, Hefei University of Technology (HFUT), Hefei 230601, China, and also are with the Institute of Artificial Intelligence, Hefei Comprehensive National Science Center, Hefei 230026, China (e-mail: eric.mengwang@gmail.com).}
}

\markboth{IEEE TRANSACTIONS ON XXX, VOL. X, 2024}%
{Shell \MakeLowercase{\textit{et al.}}: A Sample Article Using IEEEtran.cls for IEEE Journals}

 
\maketitle

\begin{abstract}
Mental health has attracted substantial attention in recent years and LLM can be an effective technology for alleviating this problem owing to its capability in text understanding and dialogue. However, existing research in this domain often suffers from limitations, such as training on datasets lacking crucial prior knowledge and evidence, and the absence of comprehensive evaluation methods. In this paper, we propose a specialized psychological large language model (LLM), named PsycoLLM, trained on a proposed high-quality psychological dataset, including single-turn QA, multi-turn dialogues and knowledge-based QA. Specifically, we construct multi-turn dialogues through a three-step pipeline comprising multi-turn QA generation, evidence judgment, and dialogue refinement. We augment this process with real-world psychological case backgrounds extracted from online platforms, enhancing the relevance and applicability of the generated data. Additionally, to compare the performance of PsycoLLM with other LLMs, we develop a comprehensive psychological benchmark based on authoritative psychological counseling examinations in China, which includes assessments of professional ethics, theoretical proficiency, and case analysis. The experimental results on the benchmark illustrate the effectiveness of PsycoLLM, which demonstrates superior performance compared to other LLMs.\footnote{https://github.com/MACLAB-HFUT/PsycoLLM}
\end{abstract}


\section{Introduction}

\IEEEPARstart{L}{arge} language models (LLM) have attracted tremendous attention and experienced promising advancements, such as ChatGPT\footnote{\url{https://chat.openai.com/}}, KimiChat\footnote{\url{https://kimi.moonshot.cn/}}, PaLM \cite{chowdhery2023palm} , BaiChuan \cite{yang2023baichuan} and LLama \cite{touvron2023llama}. 
These models have demonstrated significant advancements across a variety of tasks, including question answering (QA), summarization \cite{zhang2023summit}, and information extraction \cite{wei2023zero}. 
Moreover, \textcolor{black}{LLMs} have been effectively applied to diverse domains, such as the medical and legal domains \cite{xiong2023doctorglm,cui2023chatlaw,chen2023bianque}.
In recent years, the escalating pressures from personal, and occupational demands have emerged as significant challenges to the mental health status of individuals. 
Since LLMs are trained on massive data and contain a huge number of parameters, many experiments have shown that these models have the strong ability to understand the natural language and provide reasonable feedback.
Hence, thanks to its advanced text understanding capabilities, leveraging LLM to aid individuals in addressing psychological issues may represent an efficacious strategy for mitigating the impact of adverse mental health conditions and making it easier for seekers to obtain timely help.

Many studies have been proposed in this area.
For example, 
the work \cite{lai2023psy} combined pre-trained LLMs with real-world professional QA data, with the purpose of providing online consultation services.
\cite{chen2023soulchat} and \cite{qiu2023smile} constructed a multi-turn empathetic conversation dataset and used a base LLM for fine-tuning. 
%
\textcolor{black}{Further, EmoLLM \cite{EmoLLM} followed a similar way and fine-tuned the open base model on the mental health dataset.
}
Although these existing efforts have contributed significantly to the advancement of the field, they still exhibit certain limitations.
First, most existing psychological fine-tuning datasets are generated by even larger language models such as ChatGPT.
However, the generation process lacks adequate prior knowledge and evidence, potentially leading to the creation of datasets that deviate from real-world psychological communication scenarios.
%
%
Second, owing to the lack of a comprehensive psychological benchmark, the evaluation of most studies in this area often relies on expert assessment based on subjective judgment or overly depends on other large models, which may not fully encompass the subtle performance.
%
For instance, PsyBench \cite{zhang2023psybench} utilized questions generated by GPT-4 to evaluate model performance.

Therefore, in this paper, we propose a psychological LLM (PsycoLLM) trained on a high-quality dataset tailored to the domain of psychology, encompassing single-turn QA, multi-turn dialogues, and knowledge-based QA.
Specifically, for single-turn QA, we collect question-answer pairs from open platforms.
For multi-turn dialogues, we develop an innovative pipeline with generation, evidence judgment, and refinement for producing high-quality data.
For knowledge-based QA, we extract after-school exercises from psychology textbooks and use LLMs enhanced with agents to extract QA pairs from psychology books.
Additionally, to explore the performance of the proposed model and exiting LLMs, we introduce a benchmark based on authoritative psychological counseling examinations in China for comparative analysis.
The benchmark consists of three distinct components: professional ethics, theoretical proficiency, and case analysis.
Professional ethics and theoretical proficiency are assessed through multiple-choice questions (MCQs) to evaluate the mastery of psychological principles, while the case analysis component emphasizes practical utility by incorporating both MCQs and QA sections to explore the application of the model in real-world scenarios.

The results reveal several insightful observations:
\uppercase\expandafter{\romannumeral1}) Some LLMs have achieved an accuracy rate surpassing 60\% in MCQs, indicating their potential to excel in psychological counseling examinations, with the proposed PsycoLLM demonstrating superior performance among these models. \uppercase\expandafter{\romannumeral2}) LLMs display greater proficiency in professional ethics compared to their mastery of psychological theory.

To summarize,  the contributions of this paper are four-fold: 
\begin{itemize}
    \item We introduce PsycoLLM, a LLM fine-tuned on the high-quality psychological dataset.
    \item We present a comprehensive framework for constructing high-quality mental health datasets suitable for model training, along with the introduction of a psychological counseling examination-based benchmark for robust performance evaluation.
    \item We conduct a comparative performance analysis of various LLMs on the established benchmark. Our findings demonstrate that the proposed PsycoLLM achieves superior performance.
    
\end{itemize}

\section{Related Work}
\begin{figure*}[t]
\centering
\includegraphics[width=0.99\textwidth, trim=0 0 0 0]{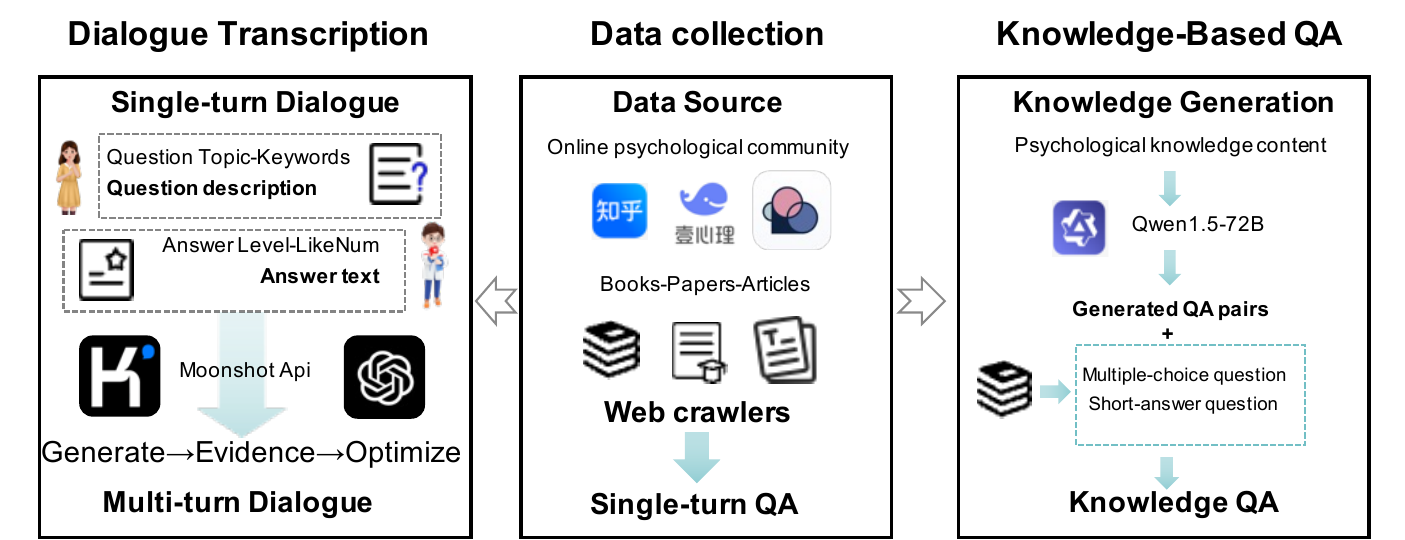}
\caption{Overview of dataset preparation, including single-turn QA, multi-turn dialogue, and knowledge QA generation.}
\label{fig:dataset_structure}
\end{figure*}

\subsection{Large Language Model}
Owing to highly parallelizable architecture of self-attention mechanisms, Transformer-based models have been popular in recent years.
BERT \cite{devlin2019bert} was a first pre-trained Transformer-based language model, which trained a bidirectional language encoder with specially designed masked language modeling.
GPT \cite{radford2018improving} trained a Transformer-based decoder by generative pre-training.
Subsequently, GPT-2 \cite{radford2019language} was introduced as an expanded version of GPT, featuring a similar architecture but with a significantly larger scale, encompassing up to 1.5 billion parameters.
BART \cite{lewis2020bart} used a sequence-to-sequence architecture and combined the strengths of BERT and GPT, which employed a bidirectional encoder similar to BERT and an autoregressive decoder similar to GPT. 
These models typically followed a pre-training and fine-tuning paradigm, which need to fine-tune these model for different downstream tasks.
Owing the findings that the larger models typically yield better performance, LLM become a prominent research topic and have achieved significant success.
For example, GPT-3 \cite{brown2020language} with 175B parameters surpassed its predecessor, GPT-2, and performed well in few-shot learning.
%

Recently, ChatGPT based on GPT-3.5, trained with a sizable amount of internet-sourced text data, has drawn sustainable attention which is exceptional at producing human-like responses.
Building upon this success, OpenAI introduced GPT-4 \cite{achiam2023gpt}, which further enhanced the capabilities of its predecessors.
In contrast to ChatGPT and GPT-4 were proprietary models, LLaMA \cite{touvron2023llama} was an open-resource language model, which attracted widespread attention from both academia and industry due to its accessible nature and has achieved noteworthy performance.
Further, Touvron et al. \cite{touvron2023llama2} proposed an updated version of LLaMA, called LLaMA2, which was trained on a larger corpus and outperformed LLaMA1 in many benchmarks.
Additionally, many other LLMs based on LLaMA architecture also have been proposed, including Alpaca \cite{taori2023alpaca} and Mistral-7B \cite{jiang2023mistral}, where the former first employed instruct-following to train the LLaMA and the latter is a 7B-parameter and outperformed LLamA-2-13B in various evaluations.
Baichuan \cite{yang2023baichuan} introduced a series of large-scale multilingual language models containing 7 billion and 13 billion parameters, trained from scratch on a corpus of 2.6 trillion tokens. 
ChatGLM \cite{glm2024chatglm} was an evolving family of large language models, which included ChatGLM \cite{du2022glm, zengglm}, ChatGLM2 and ChatGLM3, repectively.
Besides, the Yi series models \cite{young2024yi} were also trained from scratch on 3 trillion multilingual corpus and notable for their ability to extend context length up to 200K by continuing pretraining.
Furthermore, Qwen \cite{bai2023qwen} represented another open-resource LLMs and obtained superior performance across a multitude of downstream tasks, especially in Chinese, which was also the main reason we choose it as our backbone.
%

\begin{figure}[t]
\centering
\includegraphics[width=0.46\textwidth, trim=0 0 0 0]{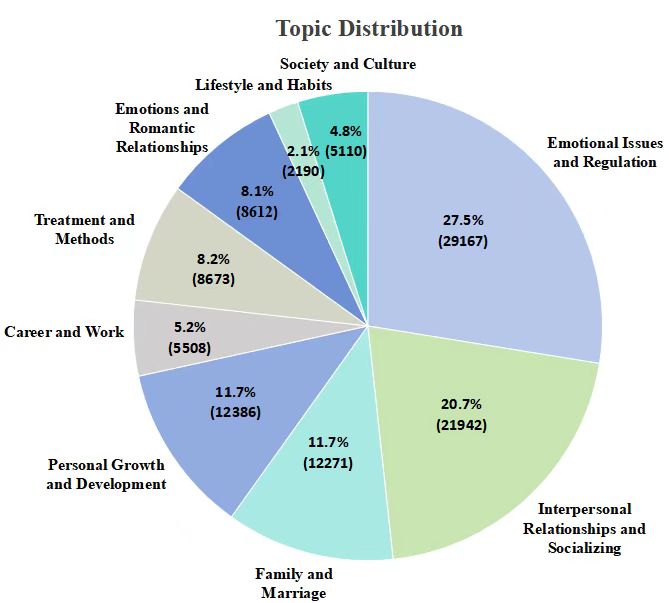}
\caption{The topic distribution. We divide the data into 9 topics and report their percentages.}
\label{fig:topic}
\end{figure}
\subsection{Benchmark for Mental Health Support}
Many studies have been devoted to evaluating the performance of the model across various domains.
For example, 
in the general domain, \cite{huang2024c} proposed a Chinese evaluation benchmark to evaluate the advanced knowledge and reasoning abilities of Chinese-based models.
%
%
\textcolor{black}{Zhong et al. \cite{zhong2024agieval} designed a human-centric benchmark for assessing the performance of foundation models via 20 diverse tasks across a wide variety of subjects.}
The MMLU benchmark \cite{hendrycksmeasuring} provided a comprehensive evaluation benchmark encompassing various tasks to evaluate the performance of a text model in multi-task contexts.
In the medical domain, \cite{wang2024cmb} proposed a localized medical benchmark across various clinical medical professions to evaluate the existing LLMs for medicine.
In the legal domain, \cite{dai2023laiw} constructed a Chinese legal benchmark based on the logic of legal practice, categorizing the legal capabilities of language models into different levels.

In the psychological domain, several studies have introduced benchmarks to assess model performance in mental health contexts.
Jin et al. \cite{jin2023psyeval} introduced a meticulously crafted benchmark to evaluate the model performance in five mental health-related sub-tasks.
Zhang et al. \cite{zhang2023psybench} used GPT-4 to generate question and constructed a psychological benchmark aimed at evaluating model performance in psychology.
Furthermore, Qiu et al. \cite{qiu2023benchmark} created a benchmark with fine-grained labels for each dialogue and focused on dialogue safety in mental health support.
Despite the advances made with these benchmarks, there are several limitations.
Some benchmarks narrow their scopes to specific subfields or isolated tasks within mental health, neglecting to evaluate the models' overall psychological knowledge mastery. Furthermore, benchmarks generated by large models carry the risk of deviating from real-world contexts.
Compared to these works, we propose a benchmark that incorporates different levels of official examinations in China, enabling a comprehensive assessment of both theoretical knowledge and practical proficiency.

%
\begin{figure}[t]
\centering
\includegraphics[width=0.48\textwidth, trim=0 0 0 0]{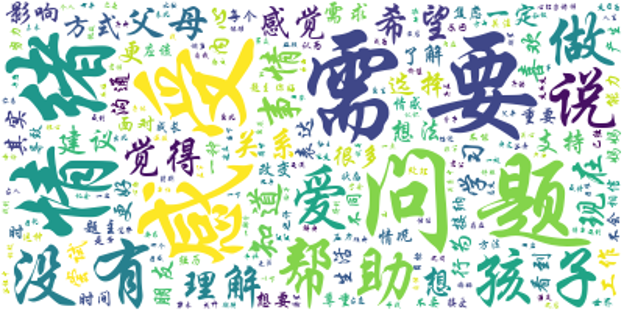}
\caption{Word cloud map of psychological consultants' response in single-turn QA dataset.}
\label{fig:word_cloud}
\end{figure}
%

\begin{figure*}[t]
\centering
\includegraphics[width=0.995\textwidth, trim=0 0 0 0]{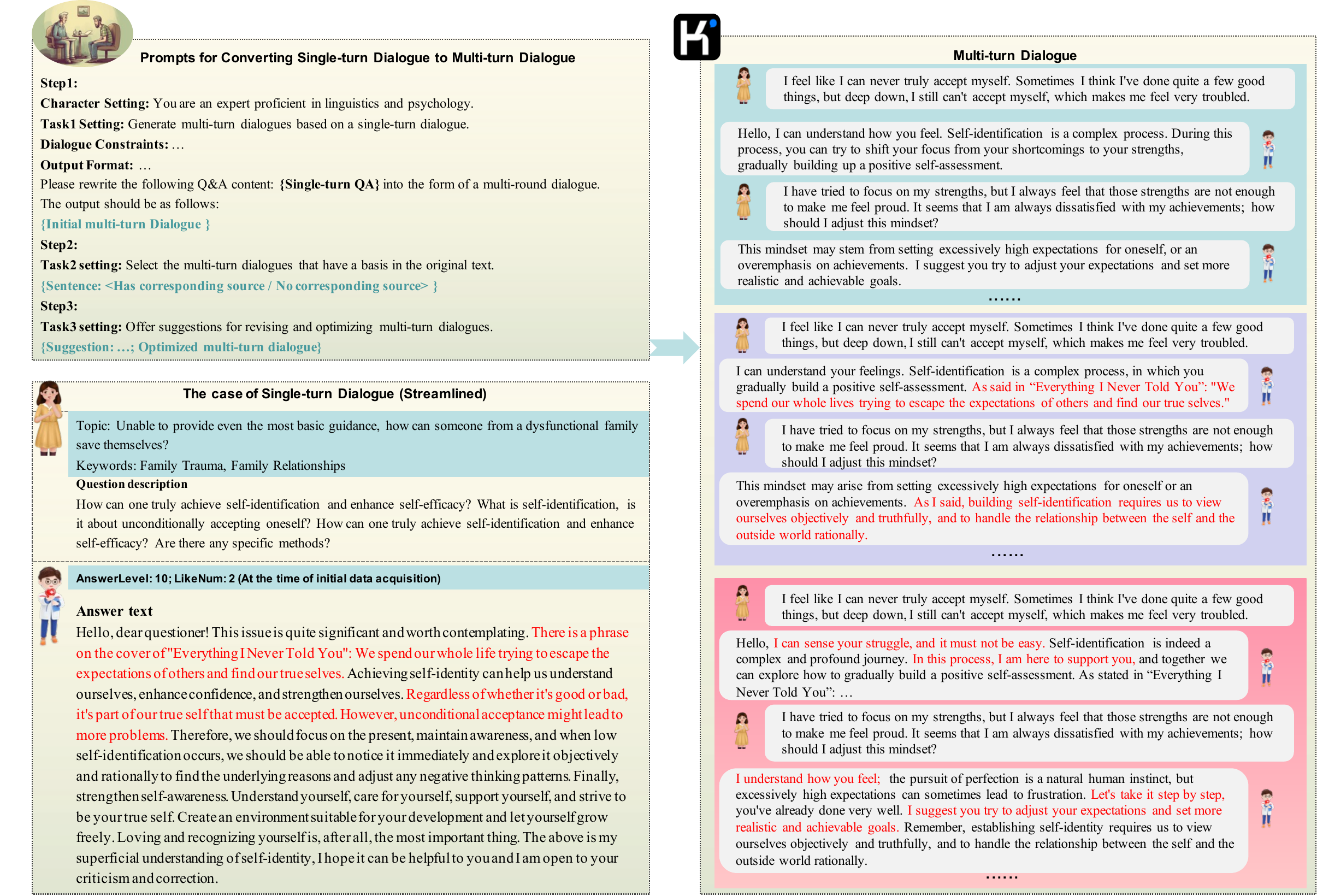}
\caption{Examples of the generated multi-turn dialogue data: Step 1 involves data generation, Step 2 focuses on evidence judgment and integration, and Step 3 entails revision for aspects such as empathy.}
\label{fig:multi_turn_case}
\end{figure*}

\subsection{Mental Application in LLM}
Many traditional deep learning-based studies have been proposed for several domains \cite{hu2021word, hu2022graph, joshi2020dr, sotudeh2022mentsum, kim2020deep,zhen2022deep}.
Recently, the success of LLMs in general domains has sparked interest in their application across various vertical fields. 
Many LLMs tailored for specific domains have achieved remarkable success \cite{huang2023lawyer, wu2024pmc, li2023chatdoctor, guo2024owl}.
For instance, HuatuoGPT \cite{zhang2023huatuogpt}, an LLM designed for the medical domain, has demonstrated excellent performance in medical consultations.
Based on HuatuoGPT, HuatuoGPT-II \cite{chen2023huatuogpt} introduced a unified domain adaptation protocol to combine continued pre-training and supervised fine-tuning into a single process, which achieved better performance than HuatuoGPT.
In the legal domain, Yue et al. \cite{yue2023disc} developed an LLM tailored for building intelligent legal systems with legal reasoning and knowledge retrieval capability. 
Mental health is another crucial domain that has garnered significant attention in many studies \cite{ilias2023calibration, hossain2024factors, ansari2022ensemble,yang2023towards,yang2024mentallama, yang2024behavioral,yang2024heterogeneous}.
For example, the work \cite{singh2024racer} developed an LLM-based expert-guided pipeline to analyze raw interview transcripts and extract insights.
Zhu et al. \cite{zhu2024reading} investigated the use of LLM to perform inferring users' underlying goals and fundamental psychological needs, finding that LLM-based approaches can be comparable to human performance.
Wu et al. \cite{wu2024sunnie} introduced an anthropomorphic LLM-based conversational agent designed to offer personalized support and suggest practical actions based on positive psychology and social psychology research.

Several LLMs have also been proposed to aid help-seekers in accessing effective mental health support.
Lai et al. \cite{lai2023psy} proposed an AI-based assistive tool that utilized LLMs for QA, aiming to alleviate the demand on mental health professionals.
PsyQA \cite{sun2021psyqa} introduced a high-quality Chinese dataset of psychological health support in the form of question and answer pairs.
%
\textcolor{black}{Based on PsyQA, Qiu et al. \cite{qiu2023smile} further used ChatGPT to convert single-turn dialogues to multi-turn dialogues and subsequently fine-tuned ChatGLM2-6B on this enhanced dataset.}
SoulChat \cite{chen2023soulchat} constructed a multi-turn empathetic conversation dataset and fine-tuned ChatGLM with this data.
Yang et al. \cite{yang2024mentallama} formalized the interpretable mental health analysis task and built a multi-task and multi-source interpretable mental health instruction dataset.
After that, based on the contructed dataset, they released MentaLLaMA, an open-source instruction-following LLM fine-tuned on LLaMA2.
The work \cite{xu2024mental} fine-tuned the LLM across multiple datasets simultaneously, demonstrating its effectiveness in various mental health prediction tasks.
Although these efforts have greatly advanced the field, they rely too heavily on distillations from other models during data generation, leading to the creation of data that may not accurately reflect real-world scenarios.
%
%
In this paper, we constructed a high-quality psychological dataset, including single-turn, multi-turn dialogue data, and knowledge QA data. 

\section{PsycoLLM}
%
\textcolor{black}{In this section, we present the details of dataset, including single-turn QA, multi-turn dialogue and knowledge-based QA, respectively.}
The overview of dataset are shown in Fig.~\ref{fig:dataset_structure}.
Besides, we give the training process of PsycoLLM.

\subsection{Single-Turn QA Construction}
\label{Yixinli}
There are several publicly accessible websites committed to establishing a psychology platform and offering online solutions for individuals seeking psychological assistance, such as Yixinli\footnote{\textcolor{black}{https://www.xinli001.com/}}, Zhihu\footnote{\textcolor{black}{https://www.zhihu.com/}} and so on.
The data from these platforms can be regarded as real-world inquiries and the solutions provided by professionals in response to these inquiries.
On the original website, users typically provide descriptions along with their questions, including background information. Subsequently, professional or experienced individuals offer detailed solutions or advice in response to these questions.
We collect over 267,000 pairs of data from websites.
Furthermore, several important data cleaning procedures are implemented to enhance the quality and relevance of the dataset, including:
\begin{itemize}
    \item Removal of irrelevant content: This involves eliminating extraneous materials such as advertisements, which do not contribute to the core data and could potentially introduce noise.
    \item Deletion of short data entries: Data entries with fewer than 100 characters are removed to ensure that the dataset contains sufficiently substantive content for meaningful analysis.
    \item Exclusion of low-engagement answers: Answers with fewer than 5 likes are excluded to prioritize content that has been deemed valuable by users. The number of likes functions as a measure of public endorsement of the response. A high number of likes suggests a broad consensus among users regarding the validity and reliability of the answer.
    \item Exclusion of responses from lower-level counselors or individual answers: This ensures the dataset reflects high-quality professional advice to some extent. The respondents' level may serve as an indicator of their extensive experience in assisting help-seekers, which, in turn, can be indicative of their professional competence. This metric provides a basis for assessing the quality of the provided information.

\end{itemize}


\begin{table*}[t]

\caption{Prompt used to generate the multi-turn dialogue.}

\centering
\resizebox{.97\textwidth}{!}{
\begin{tabular}{l|l}
\toprule[1pt]
\textsc{\textbf{Category}} & \textsc{\textbf{Prompt Template}}  \\
\midrule  

\textbf{Multi-Turn Generation}   & 
\begin{tabular}[c]{@{}l@{}}You are an expert who has studied countless conversations between patients with mental health problems and doctors... \\
Please construct a continuous multi round dialogue record between help-seekers and psychologist ...  \\
You will see a Q\&A consisting of user questions and responses from a psychological assistant. \\
The output format follows: \\
User: statement from the patient.
Psychological assistant: the doctor's comfort, advice, and guidance. \\
User:.
Psychological assistant:. \\
The dialogue should have more conversation rounds.
\end{tabular} 
\\ \hline

\textbf{Evidence Support}   & \begin{tabular}[c]{@{}l@{}}You are a professional psychologist who has provided numerous psychological consultations ... \\
Your task is to identify the evidence source of each response from the psychologist in the multi-turn dialogue Q\&A. \\
\lbrack Important \rbrack Please identify the evidence of each sentence in the multi-turn conversation from the original Q\&A ... \\
If there is no corresponding source, simply output "No corresponding source" \\
\lbrack Important \rbrack Output format: \\
response in multi-turn dialogue:\textless{}Original dialogue\textgreater \\
Source:\textless{}From visitor's self description/From doctor's reply/No corresponding source\textgreater:\textless{}Corresponding original text\textgreater
\end{tabular} 
\\ \hline

\textbf{Dialogue Evaluation}   & \begin{tabular}[c]{@{}l@{}}You are a psychology professor and are proficient in psychology and linguistics ... \\
From the perspective of a psychology professor, please make a strict judgment based on the given four indicators. \\
1. Empathy, whether the psychologist can truly understand the emotions and needs of the visitor, express sincere care ... \\
2. Supportive, whether the psychologist expresses listening to the visitor through verbal feedback ... \\
3. Guiding, whether the psychologist guides the visitor through questioning or providing choices...\\
4. Safety,  whether it may have a negative impact on visitors ...
\end{tabular} 
\\ 

\bottomrule
\end{tabular}
}
  \label{Tab:prompt}
\end{table*}
%
%
%
%
%
After these data cleaning procedures, we obtain over 155k pairs for single-turn QA.
%
We report detailed statistics of the processed data drawn from different aspects.
We divide the processed data into 9 major topics, as illustrated by the distribution of major topics depicted in Fig.~\ref{fig:topic}.
It is observed that the number of reports related to emotional issues and regulation, as well as interpersonal relationships and socializing, is higher than those in other categories, each comprising more than 20\% of the total. 
Following these topics, family and marriage, and personal growth and development are the next most significant areas, each accounting for more than 10\% of the reports.
Besides, the visualization of the collected data is depicted through word clouds, as shown in Fig.~\ref{fig:word_cloud}, which provides an intuitive depiction of the frequency and prominence of specific terms within the dataset.
%
%
The analysis underscores the pervasive connection of the majority of these data to everyday psychological counseling, further emphasizing the relevance and reasonability of the collected data.

%

\begin{table}[t]

\caption{The statistics of Multi-Turn Dialogue Data.}

\centering
\resizebox{.40\textwidth}{!}{
\begin{tabular}{l|r}
\toprule[1pt]
\multirow{1}{*}{\textsc{\textbf{\makecell[c]{Category}}}} & \textsc{\textbf{Report\#}}  \\
\midrule

\text{\# Context-Response Pairs} & 11511  \\
\text{\# Average Turns per Dialogue } & 5.90   \\
\text{\# Average Tokens per Turn } & 136.28  \\
\text{\# Average Tokens per Question} & 43.62  \\
\text{\# Average Tokens per Answer} & 92.67  \\

\bottomrule
\end{tabular}
}
  \label{Tab:multi_turn_data_statistic}
\end{table}
%
%
%
%

\begin{table*}[t]
\caption{The statistics for the benchmark dataset, where MCQ1 indicates MCQ with a single correct option, and MCQ2 denotes MCQ with multiple correct options. Levels 2 and 3 correspond to the Level 2 and Level 3 Psychological Examinations, respectively, as administered in authoritative psychological counseling examinations in China.}

\centering
\resizebox{.85\textwidth}{!}{
\begin{tabular}{l|rr|rr|rrr}
\toprule
{\multirow{2}{*}{{ \textsc{\textbf{\makecell[c]{\\Category}}}}}}& \multicolumn{2}{c|}{{Professional ethics}} & \multicolumn{2}{c|}{{Theoretical knowledge}} & \multicolumn{3}{c}{{Case Analysis}} \\

\cmidrule(r){2-8}
& \textsc{\textbf{SMCQ}} 
& \textsc{\textbf{MMCQ}} 
&\textsc{\textbf{SMCQ}} 
&\textsc{\textbf{MMCQ}}
&\textsc{\textbf{SMCQ}} 
& \textsc{\textbf{MMCQ}} 
& \textsc{\textbf{QA}}  \\
\cmidrule(lr){1-8}

\textsc{Level 2} & {48} & {48}  & {337}  & {228} & {245} & {214} & 44  \\
\textsc{Level 3} & {72} & {72}  & {566}  & {363} & {338} & {455} &40  \\
\textsc{Others} & {32} & {38}  & {241}  & {192} & {165} & {209} &16 \\
\textsc{Total} & {152} & {158}  & {1144}  & {783} & {748}  & {878} &100 \\

\bottomrule
\end{tabular}
}

\label{Tab:benchmark_statistic}
\end{table*}
\subsection{Multi-Turn Dialogue Construction}
Based on the extracted data in section \ref{Yixinli}, we select a subset of the highest upvoted answers for each question to generate multi-turn dialogue data using KimiChat.
This process aims to emulate the behavior of a psychological counselor who progressively asks probing questions to gain insight into the user's psychological state and requirements, thereby facilitating more effective assistance.
The multi-turn dialogue mainly includes two roles: the person seeking help and psychological professionals.
%
%
%
We employ a three-step pipeline to generate high-quality multi-turn dialogue data.
First, we use an appropriate prompt to guide the KimiChat to construct multi-turn dialogue data between pre-defined roles according to the selected pairs.
The original pairs serve as prior knowledge to guide KimiChat in generating multi-turn conversations, thereby enhancing the fidelity of the generated data to real-world conversations.
Second, we employ an additional prompt to assess whether the answers in the multi-turn dialogue dataset are derived from the original context.
If the majority of responses in the multi-turn dialogue can be supported by evidence extracted from the original text data, it is considered to better reflect the conversational flow of realistic counselors.
Conversely, if most responses are generated predominantly by the model without effectively leveraging the given context, such data entries require further processing.
This can be achieved by using an additional prompt to enhance the integration of factual evidence and incorporate content-relevant information from the original context without compromising fluency.
Third, to enhance the quality of the multi-turn dialogue data further, we utilize a prompt to revise the data in terms of empathy, supportiveness, guidance, and safety.
%
%
%
Ultimately, we perform manual proofreading to further ensure data quality and to derive the processed multi-turn dialogue data.
The prompts used for multi-turn dialogue generation, evidence support identification, and dialogue evaluation are presented in Table \ref{Tab:prompt}.

A example is shown in Fig.~\ref{fig:multi_turn_case}, with three steps to generate high-quality data.
In fact, we have compared using a single complex prompt to achieve the same goal as the aforementioned three steps. 
However, the utilization of a single prompt often results in an abundance of model-generated templated responses, rather than the comprehensive integration of provided factual information.
This is the primary reason we opt for a pipeline approach to generate data.
%
%
The statistics of the data is reported in Table \ref{Tab:multi_turn_data_statistic}.

\subsection{knowledge-based QA Construction}
In addition to the daily QA interactions, we also introduce some more abstract psychological knowledge data, such as explanations of psychological terms.
We crawl books related to psychology from the web and then use Qwen-72B to extract knowledge-based QA from them.
Specifically, we segment books into text spans using a predefined fixed length, identifying the nearest sentence or paragraph as segmentation indicators. 
These text spans serve as the fundamental units for subsequent QA generation through the utilization of LLMs.
First, the LLM generates questions and their corresponding answers.
These question-answer pairs are then input into two LLM-based student modules, one utilizing retrieval-augmented generation (RAG) and the other without RAG, to produce two new sets of answers.
Subsequently, a teacher module, also based on an LLM, evaluates and selects the best answer from those generated by the student modules.
Furthermore, to ensure the quality and accuracy of the generated QA pairs, a manual validation process is implemented, wherein human evaluators assess and eliminate low-quality data. 
%
In addition, we extract after-school exercises from several books along with their corresponding answer analyses.
Finally, we obtain \textcolor{black}{10K} knowledge-based QA data. 

\subsection{Supervised Fine-Tuning}
After obtaining the high-quality psychological dataset, we further conduct supervised fine-tuning (SFT) to enhance the capability of the LLM in psychology.
We select Qwen1.5-14B-Chat as our backbone model, a decoder-only architecture demonstrating excellent performance in both English and Chinese.
We then perform SFT on the top of Qwen1.5-14B-Chat to develop \textbf{PsycoLLM}.
At training stage, assuming the text input as a sequence, e.g., $x = \{x_{1}, x_{1}, . . . , x_{L}\}$, where each $x_{i}$ is a text token and $L$ is the length of $x$.
The core architecture of Qwen is based on Transformer-Decoder structure \cite{vaswani2017attention}, which is a typical autoregressive framework designed to sequentially predict each subsequent word in a given sequence. 
This can be mathematically formulated as follows:
\begin{equation}
\setlength\abovedisplayskip{6pt}
\setlength\belowdisplayskip{6pt}
    p({x})=\prod_{t=1}^{L} p\left(x_{t} \mid x_{1}, \ldots, x_{t-1}\right)    
\end{equation}
%
For training PsycoLLM, we utilize the cross-entropy loss function as the objective. The model is trained to maximize the negative conditional log-likelihood of the predicted sequence given the input data. 
\begin{equation}
\setlength\abovedisplayskip{6pt}
\setlength\belowdisplayskip{6pt}
    \theta^{*}=\underset{\theta}{\arg \max}  \sum_{t=1}^{L} \log p\left(x_{t} \mid x_{1}, ...,  x_{t-1};\theta\right),
\end{equation}
where $\theta$ is the trainable parameters of the model.
%

%
%
%
The hyperparameters utilized for training the model are configured as follows: the learning rate is set to 1e-5, the batch size is set to 128, and the number of epochs is set to 3. 
We employ full parameter fine-tuning to train the model using eight A6000 GPUs, each equipped with 48GB memory.

\section{Benchmark}
In this section, we present the psychological benchmark and provide comprehensive details, including data preparation and evaluation metrics.

\begin{figure*}[t]
\centering
\includegraphics[width=0.98\textwidth, trim=0 0 0 0]{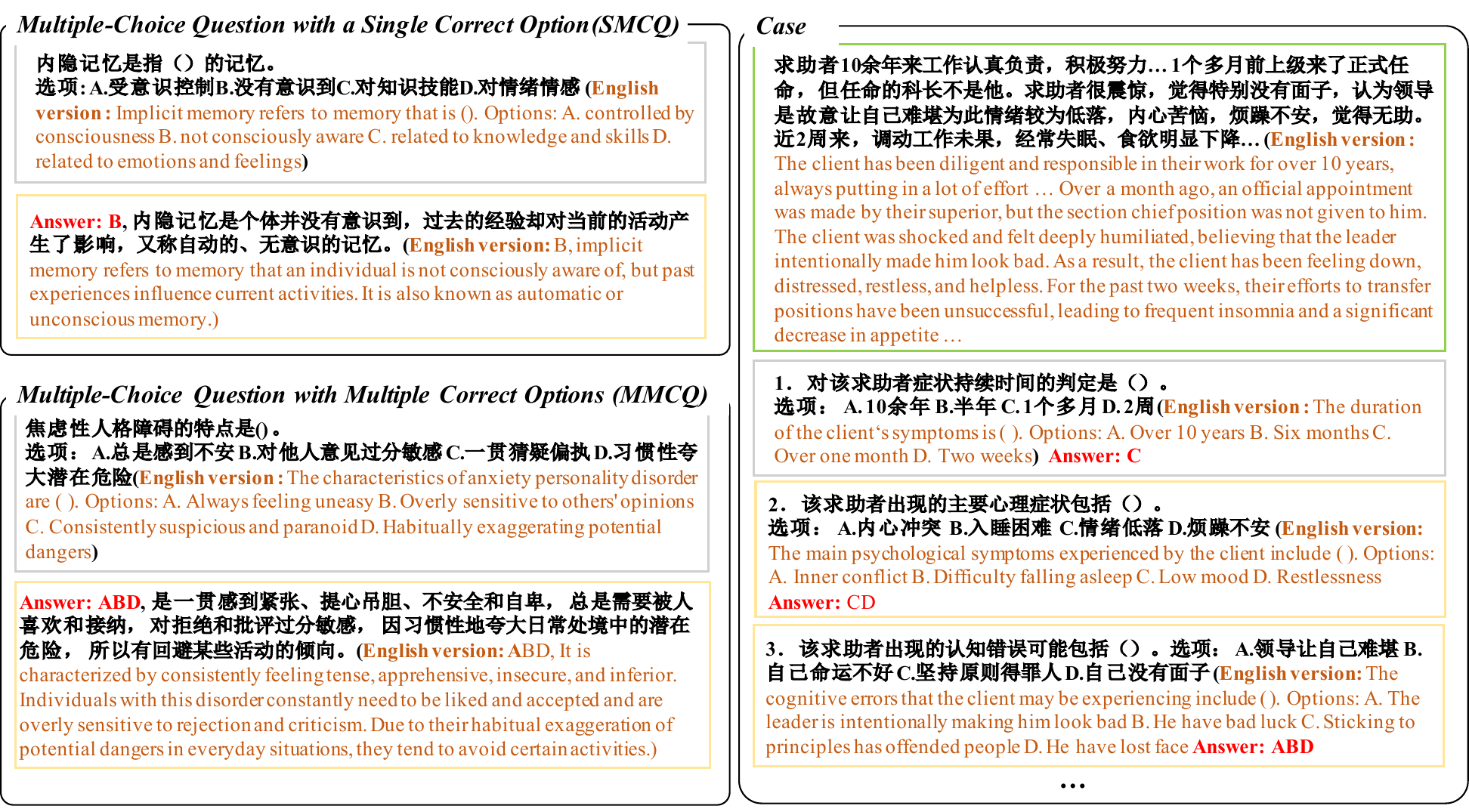}
\caption{Examples of MCQs in the proposed benchmark, including SMCQ, MMCQ and case-based MCQs, respectively.}
\label{fig:evaluation_data}
\end{figure*}
\subsection{Dataset Collection and Preprocessing}

In order to accurately evaluate the model performance in the psychology domain, we introduce a benchmark that psychological professional individuals need to master.
The data is sourced from publicly available examination questions.
We follow a standardized data preprocessing procedure.
For some data, we first need to perform optical character recognition (OCR) to convert images into text.
After that, we invite several students to manually review the collected data to ensure its quality and consistency with the original document.
This process involves rectifying formatting errors, eliminating duplicate questions, and rectifying any instances of garbled characters.
Our proposed psychological benchmark draws inspiration from the format of the most authoritative psychological counseling examination in China, comprising two primary components: objective questions and subjective questions.

Note that the objective questions mainly comprise two types of MCQ: MCQ with only a single correct option (SMCQ) and MCQ with multiple correct options (MMCQ).
The content of benchmark can be categorized into three sections: professional ethics, theoretical proficiency, and case analysis.
The professional ethics section generally encompasses general principles such as honesty, integrity, professional responsibility, and social responsibility.
Theoretical proficiency typically pertains to fundamental theories and concepts in psychology, including cognitive psychology, developmental psychology, and related fields.
Both the professional ethics and theoretical proficiency sections contain only MCQs, designed to evaluate the mastery of knowledge within the psychological domain.
For case analysis, it includes both MCQs and iterative questioning based on given case, aiming to better evaluate performance in real-world scenarios.
The former focuses on assessing the grasp of professional knowledge related to the case, while the latter offers a more open-ended assessment of the model's capability in practical utility.
%
%
%
%
Finally, we obtained a total of \textcolor{black}{3863} MCQs and \textcolor{black}{20} cases with \textcolor{black}{100} QAs.
An example of MCQs is shown in Fig. \ref{fig:evaluation_data}, and an example of case-based QA is presented in Fig. \ref{fig:case_example}.
The statistics of the benchmark are shown in Table \ref{Tab:benchmark_statistic}, detailing the distribution of questions across various categories.
The questions are classified into three main categories: Level 2 Psychological Counselor Exam, Level 3 Psychological Counselor Exam, and Others. Note that, according to the official definitions, the Level 2 exam is considered more challenging than the Level 3 exam.

\subsection{Evaluation Metric}
There are three elements in each instance of the case analysis.
$\mathcal{B}$ is the description of the given case, including the background, environment, etc. of the seeker, $\mathcal{Q}$ is the question according to the case, such as the diagnosis, reason and treatment for the case and $\mathcal{A}$ is the reference answer for each questions.
For QA in case analysis, we use the models to generate the response $\mathcal{O}$ according to the given $\mathcal{B}$ and $\mathcal{Q}$.
We evaluate the response $\mathcal{O}$ from model with the corresponding references $\mathcal{A}$.
We follow existing text generation metric, including ROUGE-1 (R-1), ROUGE-L (R-L) \cite{lin2004rouge}, BLEU-4 (B-4) \cite{papineni2002bleu}, and BERTScore \cite{zhang2019bertscore}.
For the ROUGE metric, we report the $\textsc{F}_{1}$ scores for both R-1 and R-L.
R-1 measures unigram overlap, which serves as an indicator of informativeness, while R-L evaluates the longest common subsequence overlap, providing an assessment of fluency.
BERTScore leverages embedding from pre-trained encoder to capture the semantic and contextual information of tokens through computing cosine similarities.
This metric allows for a more nuanced evaluation of text generation quality by considering the underlying meaning and context rather than just surface-level token matching.

For SMCQ, we use the standard accuracy to evaluate the model performance.
For MMCQ, we employ two types of accuracy metrics: standard accuracy and elastic accuracy.
Standard accuracy is defined as the ratio of completely correct answers to the total number of questions, providing a clear measure of the model's precision by counting only fully accurate responses.
Elastic accuracy, on the other hand, accounts for both fully accurate and partially correct answers, where latter contain no incorrect options but are not entirely complete.
This metric offers a evaluation by recognizing the model's ability to produce answers that are mostly correct, even if they are not fully comprehensive. 
The elastic accuracy for a MMCQ is computed as follows:
$$
\text{elastic accuracy} = \frac{|\text{predicted answer}|}{|\text{correct answer}|},
$$
where $||$ indicates the count of choices in the answer.

\section{Experimental Setting}

\begin{table*}[t]

\caption{Comparisons of different model on the proposed benchmark. \textsc{Ethics}, \textsc{Theory} and \textsc{Case} are professional ethics, theoretical proficiency and case analysis, respectively. The underlined number indicates the elastic accuracy for MMCQ.
The average value represents the overall average of the standard accuracy rates and the value in parentheses denotes the mean of the standard accuracy for SMCQ and the elastic accuracy for MMCQ.
}

\centering
\resizebox{.98\textwidth}{!}{
\begin{tabular}{l|ccc|ccc|ccc|c}

\toprule

 {\multirow{2}{*}{{\textsc{\textbf{\makecell[c]{Model}}}}}} & 
 
 \multicolumn{3}{c|}{\textsc{\textbf{Ethics}}} & 
 \multicolumn{3}{c|}{\textsc{\textbf{Theory}}} &
 \multicolumn{3}{c|}{\textsc{\textbf{Case}}} &
\multirow{2}{*}{{\textsc{\textbf{\makecell[c]{AvG.}}}}}
 \\

& \textsc{\textbf{SMCQ}} 
& \multicolumn{2}{|c|}{{\textsc{\textbf{MMCQ}}}}

&\textsc{\textbf{SMCQ}} 
& \multicolumn{2}{|c|}{{\textsc{\textbf{MMCQ}}}}

&\textsc{\textbf{SMCQ}} 
& \multicolumn{2}{|c|}{{\textsc{\textbf{MMCQ}}}}
\\

\cmidrule(lr){1-11}

\textsc{ChatGLM3-6B} & {63.81} & {37.97}  & \underline{52.21}  
& {46.36} & {25.08}  & \underline{36.96} 
& {37.93}  & {15.94}  & \underline{29.12}
& {37.84 (\underline{44.39})}
\\

\textsc{Yi-1.5-6B} & {71.05} & {40.50}  & \underline{54.79}
& {63.72} & {26.27}  & \underline{41.63} 
& {44.51}  & {18.90}  & \underline{27.97}
& {44.15 (\underline{50.61})}
\\

\textsc{Qwen1.5-7B-Chat} & {79.81} & {53.16}  & \underline{64.50}  
& {61.88} & {27.33}  & \underline{43.23}  
& {40.80} & {20.12}  & \underline{39.51}  
& {47.18 (\underline{54.95})} 
\\


\textsc{LLaMA3-Chinese} & {42.10}  & {18.35}  & \underline{25.84}
& {29.54}  & {13.64}  & \underline{18.68}
& {21.39}  & {10.70}  & \underline{21.92}
& {22.62 (\underline{26.57})}
\\

\textsc{Baichuan2-13B} & {64.47} & {39.87}  & \underline{50.41} 
& {52.49} & {25.69}  & \underline{37.61} 
& {37.13}  & {8.99} & \underline{29.24}
& {38.10 (\underline{45.22})}
\\

\textsc{GPT-3.5-turbo} & {74.34} & {22.78}  & \underline{43.98} 
& {55.90} & {15.01}  & \underline{36.52} 
& {49.73}  & {6.71} & \underline{33.62}
& {37.41 (\underline{49.01})}
\\
\textsc{GPT-4o} & {88.15} & {33.54}  & \underline{54.79}
& \textbf{74.65} & {24.10}  & \underline{45.07}
& \textbf{65.53}  & {13.67} & \underline{34.53}
& {49.94 (\underline{60.45})}
\\

\textsc{Qwen1.5-14B-Chat} & {80.26} & {55.69}  & \underline{67.93}  
& {70.69} & {32.44}  & \underline{49.68} 
& {41.01}  & {21.52} & \underline{41.88}
& {50.26 (\underline{58.57})}
\\

\textsc{Qwen-turbo} & {82.23} & {60.75}  & \underline{73.25}  
& {69.96} & {29.81}  & \underline{48.10} 
& {41.55}  & {24.48} & \underline{\textbf{45.11}}
& {51.38 (\underline{60.03})}
\\

\textsc{Yi-1.5-9B} &81.57 &50.63 & \underline{66.40}  
&70.51 &33.67 & \underline{50.39} 
& {50.53}  & {21.86} & \underline{41.17}
& {51.46 (\underline{60.09})} 
\\

\cmidrule(lr){1-11}
\textsc{MindChat} & {55.26} & {13.29}  & \underline{30.13} 
& {33.94} & {7.55}  & \underline{23.35} 
& {9.65}  & {4.78} & \underline{12.09}
& {20.74 (\underline{27.40})}
\\
\textsc{EmoLLM} & {55.26}  & {20.25}  & \underline{33.06}
& {38.54}  & {12.75}  & \underline{24.40} 
& {24.59}  & {11.88}  & \underline{22.28}
& {27.21 (\underline{33.02})}
\\

\cmidrule(lr){1-11}

\textsc{PsycoLLM} 
& \textbf{88.81} & \textbf{69.62}  & \textbf{\underline{74.20}}  
& {72.63} & \textbf{48.59}  & \textbf{\underline{54.12}}
& {55.58} & \textbf{35.07} & \underline{42.89} 
& \textbf{61.71 (\underline{64.70}) }\\


\bottomrule
\end{tabular}
}

\label{Tab:overall_performance}
\end{table*}

\begin{table}[t]
\caption{Comparisons of different model on the proposed benchmark, where {R-1}, {R-L}, {B-4} and {BS} are {Rouge-1}, {Rouge-L}, {Bleu-4} and {BERTScore}.
}
\centering
\resizebox{.48\textwidth}{!}{
\begin{tabular}{l|cccc}

\toprule

 {\multirow{2}{*}{{\textsc{\textbf{\makecell[c]{Model}}}}}} & 
 
 \multicolumn{4}{c}{\textsc{\textbf{Case}}} 
 \\

& \textsc{\textbf{R-1}} 
& \textsc{\textbf{R-L}} 
& \textsc{\textbf{B-4}} 
& \textsc{\textbf{BS}} 
\\

\cmidrule(lr){1-5}

\textsc{ChatGLM3} &22.00 & {14.12} & 1.82  & 64.44\\
\textsc{Baichuan2-13B} &24.06 & 17.74 & 1.93 & 64.43\\

\textsc{LLaMA3-Chinese} & {23.11} & {17.28} & {1.25} & {62.18} \\

\textsc{GPT-3.5-turbo} &22.92 & 14.19 &1.59 & 64.86 \\
\textsc{GPT-4o} & 22.09 & 15.36 & 1.39 & 64.69 \\
\textsc{Qwen1.5-14B-Chat} & 22.51 & 13.41  & 1.49 & 64.16\\
\textsc{Qwen-turbo} &21.77 &  13.21 & 1.34	& 63.97 \\
\textsc{Yi-1.5-9B} & 21.93 &  15.35	 & 1.39	& 64.51 \\
\textsc{MindChat} &24.34 &  {17.49} & 1.71 & 63.59 \\

\textsc{EmoLLM} & {23.34} & \textbf{17.98} & {1.94} & {63.41} \\

\textsc{PsycoLLM} & \textbf{24.45}  & {17.45}  & \textbf{2.04}  & \textbf{65.29}\\

\bottomrule
\end{tabular}
}

\label{Tab:subjective_performance}
\end{table}

\subsection{Baseline}
In our experiments, our experiment is implemented based on transformers\footnote{\url{https://github.com/huggingface/transformers}} and LLaMA-Factory\footnote{\url{https://github.com/hiyouga/LLaMA-Factory}}.

To explore the performance of different models, we also evaluate the following models on the proposed benchmark:
\begin{itemize}
    \item \textbf{\textsc{BaiChuan2}}: This is a generation of open-source large language models launched by Baichuan Intelligent Technology, which is trained on a high-quality corpus with 2.6 trillion tokens.
    \item \textbf{\textsc{ChatGLM3}} \cite{du2022glm}: This is a generation of pre-trained dialogue models jointly released by Zhipu AI and Tsinghua KEG.
    \item \textbf{\textsc{LLaMA3-Chinese}} \cite{Llama3-Chinese}: This model uses LLama3 as the base model and is fine-tuned with high-quality Chinese instruction datasets with the training methods of DORA and LORA+.
    \item \textbf{\textsc{Yi}} (e.g., Yi-1.5-6B, Yi-1.5-9B): These open-source large language models are trained from scratch by developers at 01.AI, which are trained with a high-quality corpus of 500B tokens.
    \item \textbf{\textsc{Qwen}} (e.g., Qwen1.5-14B-Chat, Qwen-turbo): these are decoder-only transformer models with SwiGLU activation, RoPE, multi-head attention, developed by Alibaba Cloud and are pre-trained using over 3 trillion tokens of high-quality corpus.
    \item \textbf{GPT-3.5-turbo, GPT-4o}: These models, developed by OpenAI and built upon the foundations of GPT-2 and GPT-3, excel in dialogue generation.
\end{itemize}
Besides, we compare with several existing psychological LLMs:
\begin{itemize}
    \item \textbf{\textsc{MindChat}} \cite{MindChat}\footnote{\url{https://github.com/X-D-Lab/MindChat}}: This model is fine-tuned on Qwen-7B using a mental dialogue dataset.
    \item \textbf{\textsc{EmoLLM}} \cite{EmoLLM}\footnote{\url{https://github.com/SmartFlowAI/EmoLLM}}: This model is fine-tuned on LLaMA3-8B-instruct.

\end{itemize}

\subsection{Evaluation Details}
Prompts in our experiments usually comprise following components: a well-defined task description and a pertinent question soliciting the model's response.
For tasks involving MCQ, we extract answers from model outputs using an empirically designed regular expression.

\newcolumntype{Y}{>{\centering\arraybackslash}X}

\begin{table*}[t]

\caption{Comparisons of different model on the level 2 and level 3.}

\centering
\resizebox{\textwidth}{!}{
\begin{tabular}{l|c|ccc|ccc|ccc}

\toprule
{\multirow{2}{*}{{ \textsc{\textbf{\makecell[c]{Model}}}}}} & 
{\multirow{2}{*}{{ \textsc{\textbf{\makecell[c]{Cate.}}}}}} & 
 \multicolumn{3}{c|}{\textsc{\textbf{Ethics}}} & 
 \multicolumn{3}{c|}{\textsc{\textbf{Theory}}} &
 \multicolumn{3}{c}{\textsc{\textbf{Case}}} 
  \\

&
& \multicolumn{1}{c}{\textsc{\textbf{SMCQ}}}
& \multicolumn{2}{|c|}{{\textsc{\textbf{MMCQ}}}}
& \multicolumn{1}{c}{\textsc{\textbf{SMCQ}}}
& \multicolumn{2}{|c|}{{\textsc{\textbf{MMCQ}}}}
& \multicolumn{1}{c}{\textsc{\textbf{SMCQ}}}
& \multicolumn{2}{|c}{{\textsc{\textbf{MMCQ}}}}
\\

\cmidrule(lr){1-11}

 {\multirow{2}{*}{{ \textsc{\textbf{\makecell[c]{PsycoLLM}}}}}} 
&  {Level 2} 
& 90.91 & 72.92 & \underline{77.08}
& 70.15 &47.14  & \underline{51.91} 
& 57.55 &33.64 & \underline{41.39}
\\

&  {Level 3} 
& 87.50  & 68.06 & \underline{72.11}
& 74.73  & 51.10 & \underline{56.91}
& 50.74 &36.70 & \underline{44.57}
\\

\cmidrule(lr){1-11}

 {\multirow{2}{*}{{ \textsc{\textbf{\makecell[c]{Qwen1.5-14B-Chat}}}}}}

& {Level 2} 
& 83.34  & 62.50   & \underline{70.65}
& 67.35  & 30.26  & \underline{48.83}
& 43.26 & 22.89 & \underline{40.38}
\\

& {Level 3} 
& 76.39  & 51.39 & \underline{62.27}
& 69.38  & 27.82 & \underline{47.02}
& 36.79 & 19.12 & \underline{40.54}
\\

\cmidrule(lr){1-11}

{\multirow{2}{*}{{ \textsc{\textbf{\makecell[c]{Qwen-turbo}}}}}} 

& {Level 2} 
& 87.50  & 64.58   & \underline{76.56}
& 63.79  & 32.01   & \underline{49.85}
& 43.26  & 23.83   & \underline{43.14}
  \\

& {Level 3} 
& 83.33  & 45.83   & \underline{61.80}
& 69.08  & 36.64   & \underline{53.28}
& 39.94  & 20.43   & \underline{42.01}
  \\

\bottomrule
\end{tabular}
}

\label{Tab:difficulty_performance}
\end{table*}

\section{Results and Analysis}

\subsection{Results on Objective Assessments}
To explore the performance of different models in our benchmark, we report the results of objective questions in Table \ref{Tab:overall_performance}.
There are several observations drawn from the results.
\textbf{\textit{First}}, when comparing performance between ethics and theoretical knowledge, the results indicate that these models exhibit superior performance in the ethics domain.
This discrepancy may be attributed to the closer alignment of ethics aspects with general domain dataset.
It is plausible that the training datasets for these general LLMs include a significant amount of morally relevant data, which enhances their performance in ethical evaluations. In contrast, theoretical knowledge may require more specialized data that is less prevalent in the general-domain training sets.
\textbf{\textit{Second}}, within the realm of MCQs, the accuracy for MMCQs is lower than that for SMCQs.
This difference is likely due to the increased difficulty in correctly identifying all the correct options in MMCQs.
The requirement to select multiple correct options, as opposed to a single correct option, introduces additional challenges in ensuring comprehensive and accurate responses, thereby affecting overall accuracy.
\textbf{\textit{Third}}, GPT-4o achieves the best results in SMCQ for both theoretical knowledge and case analysis.
This demonstrates that this model, despite being trained for general domain, exhibits strong performance in the psychological domain as well.
This suggests that its robust general training provides a strong foundation for effectively addressing domain-specific tasks, including those related to psychological concepts and case-based evaluations.
\textbf{\textit{Fourth}}, several models demonstrate the capability to pass the MCQs in the benchmark, primarily extracted from authoritative psychological counseling examinations.
Notably, only PsycoLLM achieves an average exceeding 60\% in standard accuracy.
In terms of elastic accuracy, GPT-4o, Qwen-turbo, Yi-1.5-9B, and PsycoLLM all achieve averages exceeding 60\%. 
This observation highlights the significant potential of using LLM for providing psychological assistance, demonstrating that these models possess the capability to pass relevant exams and deliver reliable support in this domain.
Among these models, Yi-1.5-9B is the smallest, yet it still surpasses the 60\% in elastic accuracy.
\textbf{\textit{Fifth}}, our proposed model, fine-tuned on the psychological dataset, demonstrates improvements across most evaluated aspects compared to other models, particularly in terms of standard accuracy, yielding more significant advancements.
This underscores the effectiveness of our proposed dataset in enhancing performance within the psychological domain.
\subsection{Results on Case-Based QA}
\begin{figure*}[t]
\centering
\includegraphics[width=0.97\textwidth, trim=0 0 0 0]{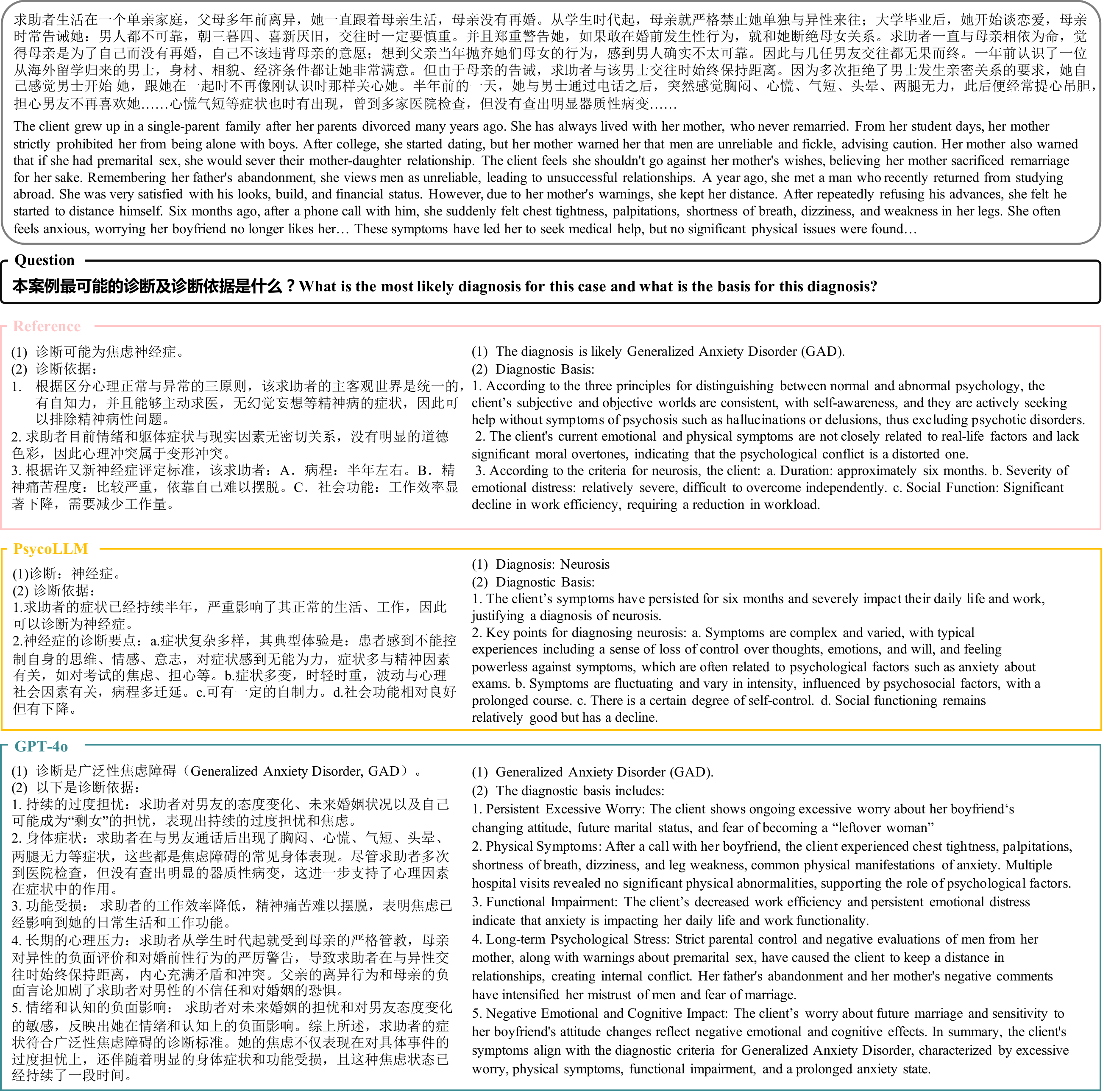}
\caption{Examples of case-based QA, including corresponding references and responses generated by PsycoLLM and GPT-4o.}
\label{fig:case_example}
\end{figure*}

We report the results of subjective questions in Table \ref{Tab:subjective_performance}.
The results obtained from automatic evaluations using ROUGE and BLEU metrics for these models are relatively low.
This may be due to the inherent limitations of these metrics in capturing the nuances and complexities of natural language generation, as they primarily focus on surface-level n-gram overlap rather than semantic equivalence and contextual appropriateness.
Therefore, we also incorporate BERTScore into our evaluation framework.
Most models consistently score between 63 and 65 on the BS indicator, indicating the consistency and stability of the BS evaluation.
\textsc{PsycoLLM} demonstrates superior performance on R-1, B-4, and BERTScore, respectively and EmoLLM achieves better results on R-L.
Additionally, although EmoLLM and MindChat do not perform well on MCQs, they achieve relatively strong performance in R-1, R-L, and B-4.
The main reason might be that these models are fine-tuned with psychological conversation data, enabling them to better grasp case information and generate higher-quality responses.
%
%
We provide a comparative example of the results from PsycoLLM and GPT-4o in Fig. \ref{fig:case_example}.
Both models are proficient at identifying the majority of patient issues.
A notable distinction emerges in their responses: GPT-4o tends to offer more detailed and specific answers, while PsycoLLM generally provides more concise responses.

\subsection{Effect of Difficulty}
We further conduct experiments to compare the results of PsycoLLM, Qwen1.5-14B and Qwen-turbo on the Level 2 and Level 3 Psychological Counselor Exams, as detailed in Table \ref{Tab:difficulty_performance}.
Among these models, the performance on theoretical proficiency is generally superior in the Level 3 examination compared to the Level 2 examination across most indicators.
This disparity may be attributed to the Level 2 examination's requirement for a more extensive understanding of psychological concepts and a deeper comprehension of the relevant theoretical frameworks.
However, it is noteworthy that these models exhibit better performance on the professional ethics and case analysis of the Level 2 exam compared to the Level 3 exam.
These observation also suggests that under certain circumstances, LLMs may perceive difficulty differently compared to humans. Not all aspects considered complex by humans necessarily pose challenges for models.

\subsection{Effect of Psychological Fine-tuning on General Ability}
\textcolor{black}{
To evaluate the impact of fine-tuning on psychological datasets with respect to general reasoning abilities, we conducted experiments comparing PsycoLLM to its non-fine-tuned baseline model, Qwen1.5-14B.
This comparison is performed across several widely used general benchmarks, including MMLU \cite{hendrycksmeasuring}, CMMLU \cite{li2023cmmlu}, GSM8K \cite{cobbe2021gsm8k} and CEVAL \cite{huang2023ceval}.
%
The results are presented in Table \ref{Tab:general_benchmark_result}.
We can find that PsycoLLM and the base model exhibit comparable performance on these general tasks, with PsycoLLM achieving even better results on the GSM8K benchmark.
This suggests that fine-tuning on the proposed psychological dataset does not significantly compromise general reasoning capabilities.
However, we observe a slight decline in several benchmarks (e.g., MMLU, CMMLU, and CEVAL), which underscores the need to balance domain-specific fine-tuning with the preservation of general reasoning abilities.
This decline may be attributed to the fact that certain tasks in these benchmarks may not been adequately considered during fine-tuning due to the differences between general and psychological domains. Additionally, there is a risk of overfitting to the psychological dataset, which could result in reduced robustness for general tasks.
}

\subsection{Discussion}
{\textit{1) Model Size:}}
On the one hand, within the same model families, models with larger parameters generally obtain better results in Table \ref{Tab:overall_performance}.
For instance, models such as YI-1.5-9B and Qwen1.5-14B-Chat demonstrate better performance compared to their smaller counterparts, YI-1.5-6B and Qwen1.5-7B-Chat, respectively.
This performance enhancement can be attributed to the increased capacity of models with larger parameters to capture complex patterns and effectively learn text representations from the data.
This observation aligns with findings from a broad spectrum of existing studies that emphasize the advantages of larger model sizes in terms of learning capacity and performance.
On the other hand, when compared models from different model families, this findings have slight difference.
For example, YI-1.5-9B outperform Baichuan2-13B and Qwen1.5-14B-Chat in the MCQ evaluation.
This discrepancy shows that training corpus, and other factors such as training procedures also significantly influence performance.

{\textit{2) Model preference in MMCQs:}}
In our experiments, we observe distinct response patterns from different models when addressing MMCQs.
For example, GPT-3.5-turbo and GPT-4o frequently provide a single option in MMCQs, whereas Qwen-turbo tends to select all available options. 
Therefore, GPT-3.5-turbo and GPT-4o have a large improvements when shifting from standard accuracy measure to elastic accuracy metric.
The main reason might be that, although GPT-3.5-turbo and GPT-4 are multilingual, they primarily focus on English. 
In English-speaking contexts, MMCQs are less prevalent compared to SMCQs.
Consequently, the training data for these models may predominantly contains SMCQs and lack a substantial MMCQ dataset. This discrepancy leads the models to exhibit a tendency to select only one answer when presented with MMCQs.

{\textit{3) Fine-tuned Models vs. General Models:}}
Among these models, EmoLLM, MindChat are fine-tuned with the psychology-related dataset and others are for general domain.
Specifically, EmoLLM and MindChat leverage empathetic conversation data for fine-tuning, which contributes to their relatively higher performance in Case-Based QA than general LLMs.
However, due to their limited incorporation of knowledge-based data, their performance in MCQ evaluations is average.

\section{Conclusion}

\textcolor{black}{
In this paper, to enhance LLM in the psychological domain, we proposed a high-quality psychological dataset.
Specifically for multi-turn dialogue, we employed a comprehensive pipeline that includes stages of generation, evidence support, and refinement. 
For knowledge-based QA, we used a teacher-student LLM-based agent to enhance data generation.
Additionally, we developed a benchmark based on psychological examinations.
The proposed PsycoLLM, fine-tuned on the psychological dataset, demonstrated superior performance on the benchmark, excelling in both knowledge mastery and case analysis compared to other LLMs.
}
\textcolor{black}{
This paper primarily focuses on the development and evaluation of a language model for psychological applications. However, it is important to note that multimodal LLMs, which incorporate both video and speech inputs, are potentially more effective for understanding user emotions due to their ability to process and integrate multiple modalities of data. The creation of multimodal datasets and benchmarks in the psychological domain warrants further investigation. We leave it for future work.
Besides, our current dataset generation process relies on LLMs, which may introduce biases inherent in these models. To address this challenge, it is crucial to explore more robust bias-mitigation strategies, which can further enhance the quality and reliability of the dataset, ultimately aiming to minimize the negative effects of bias associated with LLMs. This aspect will also be considered in future research.
}

\begin{table}[t]
\caption{Comparison of the fine-tuned model and its base model on the general benchmark.
}
\centering
\resizebox{.48\textwidth}{!}{
\begin{tabular}{l|ccccc}

\toprule

 {\multirow{2}{*}{{\textsc{\textbf{\makecell[c]{Model}}}}}} 
& \textsc{\textbf{MMLU}} 
& \textsc{\textbf{CMMLU}} 
& \textsc{\textbf{GSM8k}} 
& \textsc{\textbf{CEVAL}} 

\\
& {\textbf{5-shot}} 
& {\textbf{5-shot}} 
& {\textbf{8-shot}} 
& {\textbf{5-shot}} 
\\

\cmidrule(lr){1-5}

\textsc{Qwen1.5-14B-Chat} & 66.01  & 75.13  & 68.31 & 76.00 \\
\textsc{PsyCoLLM} & 64.55  & 74.49 & 73.39 & 75.26 \\

\bottomrule
\end{tabular}
}

\label{Tab:general_benchmark_result}
\end{table}


\bibliography{custom}

\begin{thebibliography}{10}
\providecommand{\url}[1]{#1}
\csname url@samestyle\endcsname
\providecommand{\newblock}{\relax}
\providecommand{\bibinfo}[2]{#2}
\providecommand{\BIBentrySTDinterwordspacing}{\spaceskip=0pt\relax}
\providecommand{\BIBentryALTinterwordstretchfactor}{4}
\providecommand{\BIBentryALTinterwordspacing}{\spaceskip=\fontdimen2\font plus
\BIBentryALTinterwordstretchfactor\fontdimen3\font minus \fontdimen4\font\relax}
\providecommand{\BIBforeignlanguage}[2]{{%
\expandafter\ifx\csname l@#1\endcsname\relax
\typeout{** WARNING: IEEEtran.bst: No hyphenation pattern has been}%
\typeout{** loaded for the language `#1'. Using the pattern for}%
\typeout{** the default language instead.}%
\else
\language=\csname l@#1\endcsname
\fi
#2}}
\providecommand{\BIBdecl}{\relax}
\BIBdecl

\bibitem{chowdhery2023palm}
A.~Chowdhery, S.~Narang, J.~Devlin, M.~Bosma, G.~Mishra, A.~Roberts, P.~Barham, H.~W. Chung, C.~Sutton, S.~Gehrmann \emph{et~al.}, ``Palm: Scaling language modeling with pathways,'' \emph{Journal of Machine Learning Research}, vol.~24, no. 240, pp. 1--113, 2023.

\bibitem{yang2023baichuan}
A.~Yang, B.~Xiao, B.~Wang, B.~Zhang, C.~Bian, C.~Yin, C.~Lv, D.~Pan, D.~Wang, D.~Yan \emph{et~al.}, ``Baichuan 2: Open large-scale language models,'' \emph{arXiv preprint arXiv:2309.10305}, 2023.

\bibitem{touvron2023llama}
H.~Touvron, T.~Lavril, G.~Izacard, X.~Martinet, M.-A. Lachaux, T.~Lacroix, B.~Rozi{\`e}re, N.~Goyal, E.~Hambro, F.~Azhar \emph{et~al.}, ``Llama: Open and efficient foundation language models,'' \emph{arXiv preprint arXiv:2302.13971}, 2023.

\bibitem{zhang2023summit}
H.~Zhang, X.~Liu, and J.~Zhang, ``Summit: Iterative text summarization via chatgpt,'' in \emph{Findings of the Association for Computational Linguistics: EMNLP}, 2023, pp. 10\,644--10\,657.

\bibitem{wei2023zero}
X.~Wei, X.~Cui, N.~Cheng, X.~Wang, X.~Zhang, S.~Huang, P.~Xie, J.~Xu, Y.~Chen, M.~Zhang \emph{et~al.}, ``Zero-shot information extraction via chatting with chatgpt,'' \emph{arXiv preprint arXiv:2302.10205}, 2023.

\bibitem{xiong2023doctorglm}
H.~Xiong, S.~Wang, Y.~Zhu, Z.~Zhao, Y.~Liu, L.~Huang, Q.~Wang, and D.~Shen, ``Doctorglm: Fine-tuning your chinese doctor is not a herculean task,'' \emph{arXiv preprint arXiv:2304.01097}, 2023.

\bibitem{cui2023chatlaw}
J.~Cui, Z.~Li, Y.~Yan, B.~Chen, and L.~Yuan, ``Chatlaw: Open-source legal large language model with integrated external knowledge bases,'' \emph{arXiv preprint arXiv:2306.16092}, 2023.

\bibitem{chen2023bianque}
Y.~Chen, Z.~Wang, X.~Xing, Z.~Xu, K.~Fang, J.~Wang, S.~Li, J.~Wu, Q.~Liu, X.~Xu \emph{et~al.}, ``Bianque: Balancing the questioning and suggestion ability of health llms with multi-turn health conversations polished by chatgpt,'' \emph{arXiv preprint arXiv:2310.15896}, 2023.

\bibitem{lai2023psy}
T.~Lai, Y.~Shi, Z.~Du, J.~Wu, K.~Fu, Y.~Dou, and Z.~Wang, ``Psy-llm: Scaling up global mental health psychological services with ai-based large language models,'' \emph{arXiv preprint arXiv:2307.11991}, 2023.

\bibitem{chen2023soulchat}
Y.~Chen, X.~Xing, J.~Lin, H.~Zheng, Z.~Wang, Q.~Liu, and X.~Xu, ``Soulchat: Improving llms’ empathy, listening, and comfort abilities through fine-tuning with multi-turn empathy conversations,'' in \emph{Findings of the Association for Computational Linguistics: EMNLP}, 2023, pp. 1170--1183.

\bibitem{qiu2023smile}
H.~Qiu, H.~He, S.~Zhang, A.~Li, and Z.~Lan, ``Smile: Single-turn to multi-turn inclusive language expansion via chatgpt for mental health support,'' \emph{arXiv preprint arXiv:2305.00450}, 2023.

\bibitem{EmoLLM}
\BIBentryALTinterwordspacing
EmoLLM, ``Emollm,'' 2024. [Online]. Available: \url{https://github.com/SmartFlowAI/EmoLLM/}
\BIBentrySTDinterwordspacing

\bibitem{zhang2023psybench}
J.~Zhang, H.~He, N.~Song, S.~He, H.~Qiu, A.~Li, L.~Ma, Z.~Lan \emph{et~al.}, ``Psybench: a balanced and in-depth psychological chinese evaluation benchmark for foundation models,'' \emph{arXiv preprint arXiv:2311.09861}, 2023.

\bibitem{devlin2019bert}
J.~Devlin, M.-W. Chang, K.~Lee, and K.~Toutanova, ``Bert: Pre-training of deep bidirectional transformers for language understanding,'' in \emph{Proceedings of the Conference of the North American Chapter of the Association for Computational Linguistics: Human Language Technologies}, 2019, pp. 4171--4186.

\bibitem{radford2018improving}
A.~Radford, K.~Narasimhan, T.~Salimans, I.~Sutskever \emph{et~al.}, ``Improving language understanding by generative pre-training,'' 2018.

\bibitem{radford2019language}
A.~Radford, J.~Wu, R.~Child, D.~Luan, D.~Amodei, I.~Sutskever \emph{et~al.}, ``Language models are unsupervised multitask learners,'' \emph{OpenAI blog}, vol.~1, p.~9, 2019.

\bibitem{lewis2020bart}
M.~Lewis, Y.~Liu, N.~Goyal, M.~Ghazvininejad, A.~Mohamed, O.~Levy, V.~Stoyanov, and L.~Zettlemoyer, ``Bart: Denoising sequence-to-sequence pre-training for natural language generation, translation, and comprehension,'' in \emph{Proceedings of the Annual Meeting of the Association for Computational Linguistics}, 2020, pp. 7871--7880.

\bibitem{brown2020language}
T.~Brown, B.~Mann, N.~Ryder, M.~Subbiah, J.~D. Kaplan, P.~Dhariwal, A.~Neelakantan, P.~Shyam, G.~Sastry, A.~Askell \emph{et~al.}, ``Language models are few-shot learners,'' in \emph{Advances in neural information processing systems}, 2020, pp. 1877--1901.

\bibitem{achiam2023gpt}
J.~Achiam, S.~Adler, S.~Agarwal, L.~Ahmad, I.~Akkaya, F.~L. Aleman, D.~Almeida, J.~Altenschmidt, S.~Altman, S.~Anadkat \emph{et~al.}, ``Gpt-4 technical report,'' \emph{arXiv preprint arXiv:2303.08774}, 2023.

\bibitem{touvron2023llama2}
H.~Touvron, L.~Martin, K.~Stone, P.~Albert, A.~Almahairi, Y.~Babaei, N.~Bashlykov, S.~Batra, P.~Bhargava, S.~Bhosale \emph{et~al.}, ``Llama 2: Open foundation and fine-tuned chat models,'' \emph{arXiv preprint arXiv:2307.09288}, 2023.

\bibitem{taori2023alpaca}
R.~Taori, I.~Gulrajani, T.~Zhang, Y.~Dubois, X.~Li, C.~Guestrin, P.~Liang, and T.~B. Hashimoto, ``Alpaca: A strong, replicable instruction-following model,'' \url{https://github.com/tatsu-lab/stanford alpaca}, 2023.

\bibitem{jiang2023mistral}
A.~Q. Jiang, A.~Sablayrolles, A.~Mensch, C.~Bamford, D.~S. Chaplot, D.~d.~l. Casas, F.~Bressand, G.~Lengyel, G.~Lample, L.~Saulnier \emph{et~al.}, ``Mistral 7b,'' \emph{arXiv preprint arXiv:2310.06825}, 2023.

\bibitem{glm2024chatglm}
T.~GLM, A.~Zeng, B.~Xu, B.~Wang, C.~Zhang, D.~Yin, D.~Rojas, G.~Feng, H.~Zhao, H.~Lai \emph{et~al.}, ``Chatglm: A family of large language models from glm-130b to glm-4 all tools,'' \emph{arXiv preprint arXiv:2406.12793}, 2024.

\bibitem{du2022glm}
Z.~Du, Y.~Qian, X.~Liu, M.~Ding, J.~Qiu, Z.~Yang, and J.~Tang, ``Glm: General language model pretraining with autoregressive blank infilling,'' in \emph{Proceedings of the Annual Meeting of the Association for Computational Linguistics}, 2022, pp. 320--335.

\bibitem{zengglm}
A.~Zeng, X.~Liu, Z.~Du, Z.~Wang, H.~Lai, M.~Ding, Z.~Yang, Y.~Xu, W.~Zheng, X.~Xia \emph{et~al.}, ``Glm-130b: An open bilingual pre-trained model,'' in \emph{International Conference on Learning Representations}, 2023.

\bibitem{young2024yi}
A.~Young, B.~Chen, C.~Li, C.~Huang, G.~Zhang, G.~Zhang, H.~Li, J.~Zhu, J.~Chen, J.~Chang \emph{et~al.}, ``Yi: Open foundation models by 01. ai,'' \emph{arXiv preprint arXiv:2403.04652}, 2024.

\bibitem{bai2023qwen}
J.~Bai, S.~Bai, Y.~Chu, Z.~Cui, K.~Dang, X.~Deng, Y.~Fan, W.~Ge, Y.~Han, F.~Huang \emph{et~al.}, ``Qwen technical report,'' \emph{arXiv preprint arXiv:2309.16609}, 2023.

\bibitem{huang2024c}
Y.~Huang, Y.~Bai, Z.~Zhu, J.~Zhang, J.~Zhang, T.~Su, J.~Liu, C.~Lv, Y.~Zhang, Y.~Fu \emph{et~al.}, ``C-eval: A multi-level multi-discipline chinese evaluation suite for foundation models,'' in \emph{Thirty-seventh Conference on Neural Information Processing Systems Datasets and Benchmarks Track}, 2024.

\bibitem{zhong2024agieval}
W.~Zhong, R.~Cui, Y.~Guo, Y.~Liang, S.~Lu, Y.~Wang, A.~Saied, W.~Chen, and N.~Duan, ``Agieval: A human-centric benchmark for evaluating foundation models,'' in \emph{Findings of the Association for Computational Linguistics: NAACL}, 2024, pp. 2299--2314.

\bibitem{hendrycksmeasuring}
D.~Hendrycks, C.~Burns, S.~Basart, A.~Zou, M.~Mazeika, D.~Song, and J.~Steinhardt, ``Measuring massive multitask language understanding,'' in \emph{International Conference on Learning Representations}.

\bibitem{wang2024cmb}
X.~Wang, G.~Chen, S.~Dingjie, Z.~Zhiyi, Z.~Chen, Q.~Xiao, J.~Chen, F.~Jiang, J.~Li, X.~Wan \emph{et~al.}, ``Cmb: A comprehensive medical benchmark in chinese,'' in \emph{Proceedings of the Conference of the North American Chapter of the Association for Computational Linguistics: Human Language Technologies}, 2024, pp. 6184--6205.

\bibitem{dai2023laiw}
Y.~Dai, D.~Feng, J.~Huang, H.~Jia, Q.~Xie, Y.~Zhang, W.~Han, W.~Tian, and H.~Wang, ``Laiw: A chinese legal large language models benchmark (a technical report),'' \emph{arXiv preprint arXiv:2310.05620}, 2023.

\bibitem{jin2023psyeval}
H.~Jin, S.~Chen, M.~Wu, and K.~Q. Zhu, ``Psyeval: A comprehensive large language model evaluation benchmark for mental health,'' \emph{arXiv preprint arXiv:2311.09189}, 2023.

\bibitem{qiu2023benchmark}
H.~Qiu, T.~Zhao, A.~Li, S.~Zhang, H.~He, and Z.~Lan, ``A benchmark for understanding dialogue safety in mental health support,'' in \emph{CCF International Conference on Natural Language Processing and Chinese Computing}, 2023, pp. 1--13.

\bibitem{hu2021word}
J.~Hu, J.~Li, Z.~Chen, Y.~Shen, Y.~Song, X.~Wan, and T.-H. Chang, ``Word graph guided summarization for radiology findings,'' in \emph{Findings of the Association for Computational Linguistics: ACL-IJCNLP 2021}, 2021, pp. 4980--4990.

\bibitem{hu2022graph}
J.~Hu, Z.~Li, Z.~Chen, Z.~Li, X.~Wan, and T.-H. Chang, ``Graph enhanced contrastive learning for radiology findings summarization,'' in \emph{Proceedings of the 60th Annual Meeting of the Association for Computational Linguistics (Volume 1: Long Papers)}, 2022, pp. 4677--4688.

\bibitem{joshi2020dr}
A.~Joshi, N.~Katariya, X.~Amatriain, and A.~Kannan, ``Dr. summarize: Global summarization of medical dialogue by exploiting local structures.'' in \emph{Findings of the Association for Computational Linguistics: EMNLP 2020}, 2020, pp. 3755--3763.

\bibitem{sotudeh2022mentsum}
S.~Sotudeh, N.~Goharian, and Z.~Young, ``Mentsum: A resource for exploring summarization of mental health online posts,'' in \emph{Proceedings of the thirteenth language resources and evaluation conference}, 2022, pp. 2682--2692.

\bibitem{kim2020deep}
J.~Kim, J.~Lee, E.~Park, and J.~Han, ``A deep learning model for detecting mental illness from user content on social media,'' \emph{Scientific reports}, vol.~10, no.~1, p. 11846, 2020.

\bibitem{zhen2022deep}
Z.~Zhen, R.~Wang, and W.~Zhu, ``A deep learning based method for intelligent detection of seafarers' mental health condition,'' \emph{Scientific Reports}, vol.~12, no.~1, p. 7890, 2022.

\bibitem{huang2023lawyer}
Q.~Huang, M.~Tao, C.~Zhang, Z.~An, C.~Jiang, Z.~Chen, Z.~Wu, and Y.~Feng, ``Lawyer llama technical report,'' \emph{arXiv preprint arXiv:2305.15062}, 2023.

\bibitem{wu2024pmc}
C.~Wu, W.~Lin, X.~Zhang, Y.~Zhang, W.~Xie, and Y.~Wang, ``Pmc-llama: toward building open-source language models for medicine,'' \emph{Journal of the American Medical Informatics Association}, p. ocae045, 2024.

\bibitem{li2023chatdoctor}
Y.~Li, Z.~Li, K.~Zhang, R.~Dan, S.~Jiang, and Y.~Zhang, ``Chatdoctor: A medical chat model fine-tuned on a large language model meta-ai (llama) using medical domain knowledge,'' \emph{Cureus}, vol.~15, no.~6, 2023.

\bibitem{guo2024owl}
H.~Guo, J.~Yang, J.~Liu, L.~Yang, L.~Chai, J.~Bai, J.~Peng, X.~Hu, C.~Chen, D.~Zhang, xu~Shi, T.~Zheng, liangfan zheng, B.~Zhang, K.~Xu, and Z.~Li, ``{OWL}: A large language model for {IT} operations,'' in \emph{International Conference on Learning Representations}, 2024.

\bibitem{zhang2023huatuogpt}
H.~Zhang, J.~Chen, F.~Jiang, F.~Yu, Z.~Chen, G.~Chen, J.~Li, X.~Wu, Z.~Zhiyi, Q.~Xiao \emph{et~al.}, ``Huatuogpt, towards taming language model to be a doctor,'' in \emph{Findings of the Association for Computational Linguistics: EMNLP 2023}, 2023, pp. 10\,859--10\,885.

\bibitem{chen2023huatuogpt}
J.~Chen, X.~Wang, A.~Gao, F.~Jiang, S.~Chen, H.~Zhang, D.~Song, W.~Xie, C.~Kong, J.~Li \emph{et~al.}, ``Huatuogpt-ii, one-stage training for medical adaption of llms,'' \emph{arXiv preprint arXiv:2311.09774}, 2023.

\bibitem{yue2023disc}
S.~Yue, W.~Chen, S.~Wang, B.~Li, C.~Shen, S.~Liu, Y.~Zhou, Y.~Xiao, S.~Yun, W.~Lin \emph{et~al.}, ``Disc-lawllm: Fine-tuning large language models for intelligent legal services,'' \emph{arXiv preprint arXiv:2309.11325}, 2023.

\bibitem{ilias2023calibration}
L.~Ilias, S.~Mouzakitis, and D.~Askounis, ``Calibration of transformer-based models for identifying stress and depression in social media,'' \emph{IEEE Transactions on Computational Social Systems}, vol.~11, no.~2, pp. 1979--1990, 2023.

\bibitem{hossain2024factors}
A.~Hossain \emph{et~al.}, ``Factors influencing mental health among youth during the covid-19 lockdown: A cross-sectional study in bangladesh,'' \emph{IEEE Transactions on Computational Social Systems}, 2024.

\bibitem{ansari2022ensemble}
L.~Ansari, S.~Ji, Q.~Chen, and E.~Cambria, ``Ensemble hybrid learning methods for automated depression detection,'' \emph{IEEE transactions on computational social systems}, vol.~10, no.~1, pp. 211--219, 2022.

\bibitem{yang2023towards}
K.~Yang, S.~Ji, T.~Zhang, Q.~Xie, Z.~Kuang, and S.~Ananiadou, ``Towards interpretable mental health analysis with large language models,'' in \emph{Proceedings of the Conference on Empirical Methods in Natural Language Processing}, 2023, pp. 6056--6077.

\bibitem{yang2024mentallama}
K.~Yang, T.~Zhang, Z.~Kuang, Q.~Xie, J.~Huang, and S.~Ananiadou, ``Mentallama: interpretable mental health analysis on social media with large language models,'' in \emph{Proceedings of the ACM on Web Conference 2024}, 2024, pp. 4489--4500.

\bibitem{yang2024behavioral}
M.~Yang, Y.~Tao, H.~Cai, and B.~Hu, ``Behavioral information feedback with large language models for mental disorders: Perspectives and insights,'' \emph{IEEE Transactions on Computational Social Systems}, vol.~11, no.~3, pp. 3026--3044, 2024.

\bibitem{yang2024heterogeneous}
M.~Yang, Z.~Li, Y.~Gao, C.~He, F.~Huang, and W.~Chen, ``Heterogeneous graph attention networks for depression identification by campus cyber-activity patterns,'' \emph{IEEE Transactions on Computational Social Systems}, 2024.

\bibitem{singh2024racer}
S.~H. Singh, K.~Jiang, K.~Bhasin, A.~Sabharwal, N.~Moukaddam, and A.~B. Patel, ``Racer: An llm-powered methodology for scalable analysis of semi-structured mental health interviews,'' \emph{arXiv preprint arXiv:2402.02656}, 2024.

\bibitem{zhu2024reading}
Q.~Zhu, L.~Chong, M.~Yang, and J.~Luo, ``Reading users' minds from what they say: An investigation into llm-based empathic mental inference,'' \emph{arXiv preprint arXiv:2403.13301}, 2024.

\bibitem{wu2024sunnie}
S.~Wu, F.~Han, B.~Yao, T.~Xie, X.~Zhao, and D.~Wang, ``Sunnie: An anthropomorphic llm-based conversational agent for mental well-being activity recommendation,'' \emph{arXiv preprint arXiv:2405.13803}, 2024.

\bibitem{sun2021psyqa}
H.~Sun, Z.~Lin, C.~Zheng, S.~Liu, and M.~Huang, ``Psyqa: A chinese dataset for generating long counseling text for mental health support,'' in \emph{Findings of the Association for Computational Linguistics: ACL-IJCNLP}, 2021, pp. 1489--1503.

\bibitem{xu2024mental}
X.~Xu, B.~Yao, Y.~Dong, S.~Gabriel, H.~Yu, J.~Hendler, M.~Ghassemi, A.~K. Dey, and D.~Wang, ``Mental-llm: Leveraging large language models for mental health prediction via online text data,'' \emph{Proceedings of the ACM on Interactive, Mobile, Wearable and Ubiquitous Technologies}, vol.~8, no.~1, pp. 1--32, 2024.

\bibitem{vaswani2017attention}
A.~Vaswani, N.~Shazeer, N.~Parmar, J.~Uszkoreit, L.~Jones, A.~N. Gomez, {\L}.~Kaiser, and I.~Polosukhin, ``Attention is all you need,'' in \emph{Advances in neural information processing systems}, 2017.

\bibitem{lin2004rouge}
C.-Y. Lin, ``Rouge: A {R}ackage for {A}utomatic {E}valuation of {S}ummaries,'' in \emph{Text summarization branches out}, 2004, pp. 74--81.

\bibitem{papineni2002bleu}
K.~Papineni, S.~Roukos, T.~Ward, and W.-J. Zhu, ``Bleu: a method for automatic evaluation of machine translation,'' in \emph{Proceedings of the Annual Meeting of the Association for Computational Linguistics}, 2002, pp. 311--318.

\bibitem{zhang2019bertscore}
T.~Zhang, V.~Kishore, F.~Wu, K.~Q. Weinberger, and Y.~Artzi, ``Bertscore: Evaluating text generation with bert,'' in \emph{International Conference on Learning Representations}, 2019.

\bibitem{Llama3-Chinese}
L.~C. Zhichen~Zhang, Xin~LU, ``Llama3-chinese,'' \url{https://github.com/seanzhang-zhichen/llama3-chinese}, 2024.

\bibitem{MindChat}
D.~X. Xin~Yan, ``Mindchat: Psychological large language model,'' \url{https://github.com/X-D-Lab/MindChat}, 2023.

\bibitem{li2023cmmlu}
H.~Li, Y.~Zhang, F.~Koto, Y.~Yang, H.~Zhao, Y.~Gong, N.~Duan, and T.~Baldwin, ``Cmmlu: Measuring massive multitask language understanding in chinese,'' \emph{arXiv preprint arXiv:2306.09212}, 2023.

\bibitem{cobbe2021gsm8k}
K.~Cobbe, V.~Kosaraju, M.~Bavarian, M.~Chen, H.~Jun, L.~Kaiser, M.~Plappert, J.~Tworek, J.~Hilton, R.~Nakano, C.~Hesse, and J.~Schulman, ``Training verifiers to solve math word problems,'' \emph{arXiv preprint arXiv:2110.14168}, 2021.

\bibitem{huang2023ceval}
Y.~Huang, Y.~Bai, Z.~Zhu, J.~Zhang, J.~Zhang, T.~Su, J.~Liu, C.~Lv, Y.~Zhang, J.~Lei, Y.~Fu, M.~Sun, and J.~He, ``C-eval: A multi-level multi-discipline chinese evaluation suite for foundation models,'' in \emph{Advances in Neural Information Processing Systems}, 2023.

\end{thebibliography}
\bibliographystyle{IEEEtran}


\vfill

\end{document}